 \documentclass[letterpaper, 10 pt, conference]{ieeeconf}
 \IEEEoverridecommandlockouts
\usepackage[bookmarks=true,hidelinks]{hyperref}
\usepackage[utf8]{inputenc}
\usepackage{color}
\usepackage[dvipsnames]{xcolor}
\usepackage{soul}
\usepackage{float}
\usepackage{graphicx}
\graphicspath{ {figures/} }

\usepackage{cite}
\usepackage{tikz}
\usepackage{amssymb}
\usepackage{amsmath}
\usepackage{physics}
\usepackage{breqn}
\usepackage{overpic}
\usepackage{resizegather}
\usepackage{siunitx}
\usepackage{afterpage}
\usepackage{balance}
\usepackage{textcomp}
\usepackage[export]{adjustbox} 
\usepackage{bm}
\definecolor{myDarkGrey}{RGB}{65, 65, 65}

\DeclareMathOperator{\arctantwo}{arctan2}

\begin{document}
\author{Victor M. Baez$^{1}$, Nikhil Navkar$^{2}$, Aaron T. Becker$^{1}$}
\title{\LARGE\bf An Analytic Solution to the 3D CSC Dubins Path Problem
\thanks{
This work was initiated at the \href{https://www.dagstuhl.de/23091}{Dagstuhl Seminar 23091, ``Algorithmic Foundations of Programmable Matter'' in 2023}.
     We thank Cynthia Sung for presenting this open problem and for discussions on practical applications. 
Support by the National Priority Research Program (NPRP) award (NPRP13S-0116-200084) from the Qatar National Research Fund (a member of The Qatar Foundation), 
the \href{https://www.humboldt-foundation.de/en/}{Alexander von Humboldt Foundation}, and the National Science Foundation under 
\href{http://nsf.gov/awardsearch/showAward?AWD_ID=1553063}{[IIS-1553063},
\href{https://nsf.gov/awardsearch/showAward?AWD_ID=1849303}{1849303},
\href{https://nsf.gov/awardsearch/showAward?AWD_ID=2130793}{2130793]}.
}
\thanks{
$^{1}$Electrical Engineering, University of Houston, TX USA\newline {\tt\small \{vjmontan,atbecker\}@uh.edu}}
\thanks{
$^{2}$Department of Surgery, Hamad Medical Corporation, Doha, Qatar {\tt\small NNavkar@hamad.qa}}
}%
\maketitle
\begin{abstract}
We present an analytic solution to the 3D Dubins path problem for paths composed of an initial circular arc, a straight component, and a final circular arc. These are commonly called CSC paths. 
By modeling the start and goal configurations of the path as the base frame and final frame of an RRPRR manipulator, we treat this as an inverse kinematics problem. The kinematic features of the 3D Dubins path are built into the constraints of our manipulator model. 
Furthermore,
we show that the number of solutions is not constant,
with up to seven valid CSC path solutions even in non-singular regions. 
An implementation of solution is available at \url{https://github.com/aabecker/dubins3D}.
\end{abstract}

\section{Introduction and Related Work}

In 1957, Lester Eli Dubins proved that the shortest path between two ($x,y,\theta$) coordinates for a forward-moving vehicle with a minimum turning radius $r$  is composed entirely of no more than three components, where each component is either a circular arc at minimum turning radius or a straight line~\cite{dubins1957curves}.
In a common shorthand notation, `C' is a circular arc at the maximum curvature and `S' is a straight-line component.
The shortest curvature-constrained path is fully solved in 2D, with four to six candidate solutions of the form CSC or CCC everywhere. 
The 3D extension has applications in drone control, underwater vehicles, tendon transmission paths, directional well trajectory planning, and manipulator design\cite{hota2010CDC,chen2022kinegami,vana2020,yuwang2015,anderson2011stochastic,Owen2015,Two3DDubinsVehicles2012,li2023optimization,cui2018smooth,liu2022subsea,herynek2021finding,lim2023circling,xu2023shunted,moon2023time,blevins2023real,bashi2024developing,consonni2023new,wu2023adaptive,hague2023planning,liu2023practical,patsko2023three,tian2024multi}.
In exciting recent work, Chen et al.\ in~\cite{chen2022kinegami} used the  path computation from~\cite{hota2010optimal} to plan 3D Dubins paths for robot models.

Hota and Ghose provided a numerical method to efficiently find a 3D CSC path~\cite{hota2010CDC}, and further expanded this method to find up to four different CSC paths in~\cite{hota2010optimal}.
Some works split the 3D problem into a 2D Dubins path in $(x,y)$ and a separate pitch command for $Z$, incorporating an iterative procedure as in~\cite{vana2020}, or adding a separate helical trajectory for altitude adjustment as in~\cite{yuwang2015}. 
Exhaustive search methods based on optimal control were formed and tested on the 3D Dubins path in~\cite{Wang2021}. 
However, these methods are only resolution complete and the results are only close to optimal over a search space. In later work~\cite{Wang2022}, authors of~\cite{Wang2021} describe their search as usually slow.
In contrast, this paper provides an analytic method to determine all candidate CSC paths in 3D.  Given these paths, it is then trivial to choose the shortest path, or to choose between the candidate options to satisfy  other control objectives.

\begin{figure}[t]
\centering
\includegraphics[height=0.51\columnwidth,trim={.1cm 0 1.75cm 0},clip]
{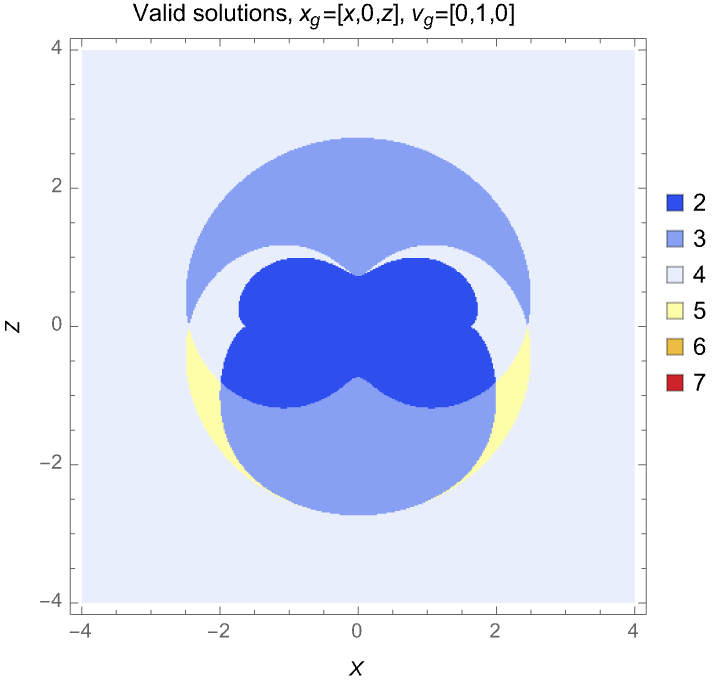}
\includegraphics[height=0.51\columnwidth,trim={.7cm 0 0.28cm 0},clip]
{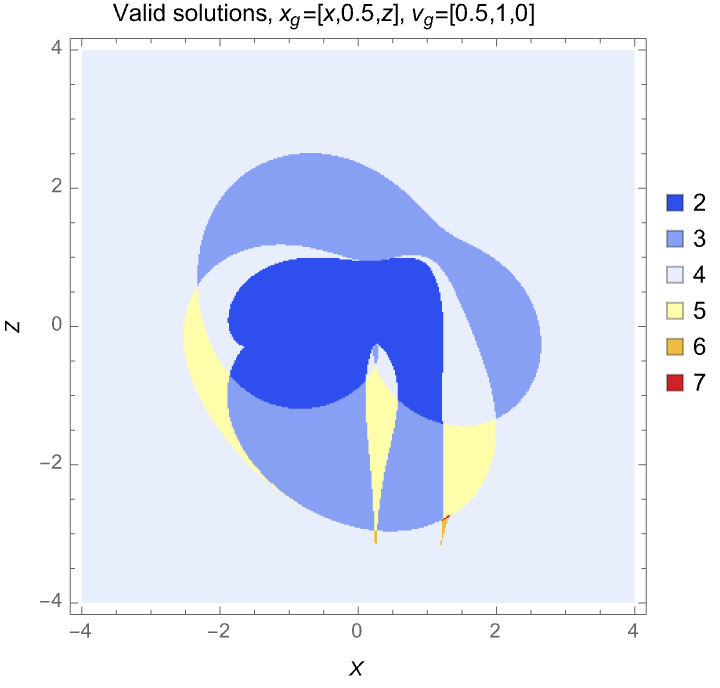}
\caption{
2D slices of the 5-DOF configuration space, showing that the number of valid solutions to the CSC Dubins path ranges from 2 to 7 for the configurations shown. See video overview at \href{https://youtu.be/aWfmgsal0JU}{https://youtu.be/aWfmgsal0JU}.}
\label{fig:2DsliceNumberSols}
\end{figure}

A key advantage of the analytic solution over a numeric approach is that it provides all possible solutions. 
As shown in Fig.~\ref{fig:2DsliceNumberSols}, the number of valid CSC solutions varies over the workspace. 
Another benefit is consistent run time. 
The numerical approach in \cite{hota2010optimal} 
took $11.0\pm11.6$ seconds 
to find all seven solutions in Fig.~\ref{fig:7SolutionExample1}, but our Mathematica implementation solves for all seven  in $0.02499\pm0.00097$ seconds.
All tests were run on a Macbook Pro with an Apple M2 Pro processor and 32 Gb unified memory. 
Converting the code to a compiled language is left for future work, but should increase the speed.

\subsection{Problem statement 
}
The CSC 3D path with a minimum curvature radius $r$ has five inputs as shown in Fig.~\ref{fig:DHtoDubins}: the bending angle of the first C path $\psi_1$, the orientation  $\phi_1$ of this angle with regard to the $(x_0,y_0)$ plane, the S extension length $d \ge 0$, the  bending angle of the second C path $\psi_2$ and the orientation $\phi_2$ of this angle with regard to the $(x_1,y_1)$ plane,
We define our coordinate plane so that the path starts at $\textbf{x}_0=[0,0,0]^\top$ with orientation $\textbf{v}_0 = [0,0,1]^\top.$
We adopt the shorthand $\cos{(\psi)} = c_\psi$, and $\sin{(\psi)} = s_\psi$. 
Then the final orientation $\textbf{v}_g$ and position $\textbf{x}_g$ are given as follows.
\begin{align}
\textbf{v}_g &= \left[\begin{array}{r}
s_{\psi_2} (c_{\psi_1} c_{\phi_1} c_{\phi_2}-s_{\phi_1} s_{\phi_2})+s_{\psi_1} c_{\psi_2} c_{\phi_1}
\\
s_{\psi_2} (c_{\psi_1} s_{\phi_1} c_{\phi_2}+c_{\phi_1} s_{\phi_2})+s_{\psi_1} c_{\psi_2} s_{\phi_1}
\\
c_{\psi_1} c_{\psi_2}-s_{\psi_1} s_{\psi_2} c_{\phi_2} 
\end{array}\right]
\\
\textbf{x}_g &= \!\resizebox{0.92\columnwidth}{!}{ $\left[\begin{array}{l} c_{\phi_1} (s_{\psi_1} (d+r s_{\psi_2})+r c_{\psi_1} (c_{\phi_2}(1-c_{\psi_2})-1)+r)+r (c_{\psi_2}-1) s_{\phi_1} s_{\phi_2}
\\
s_{\phi_1} (s_{\psi_1} (d+r s_{\psi_2})+r c_{\psi_1} (c_{\phi_2}(1 -c_{\psi_2} )-1)+r) -r (c_{\psi_2}-1) c_{\phi_1} s_{\phi_2}
\\
~\,~~~~c_{\psi_1} (d+r s_{\psi_2})+r s_{\psi_1} (c_{\phi_2}(c_{\psi_2}-1) +1)
\end{array}\!\!\right]$} \label{eq:fwdKine}
\end{align}
The inverse kinematics problem is to  solve for the $[\phi_1,\psi_1,d,\phi_2,\psi_2]$ that produce the orientation specified by a unit vector $\textbf{v}_g$ and position specified by $\textbf{x}_g$.


\section{Model}

\begin{figure}[tb]
\centering
\includegraphics[width=1\columnwidth]{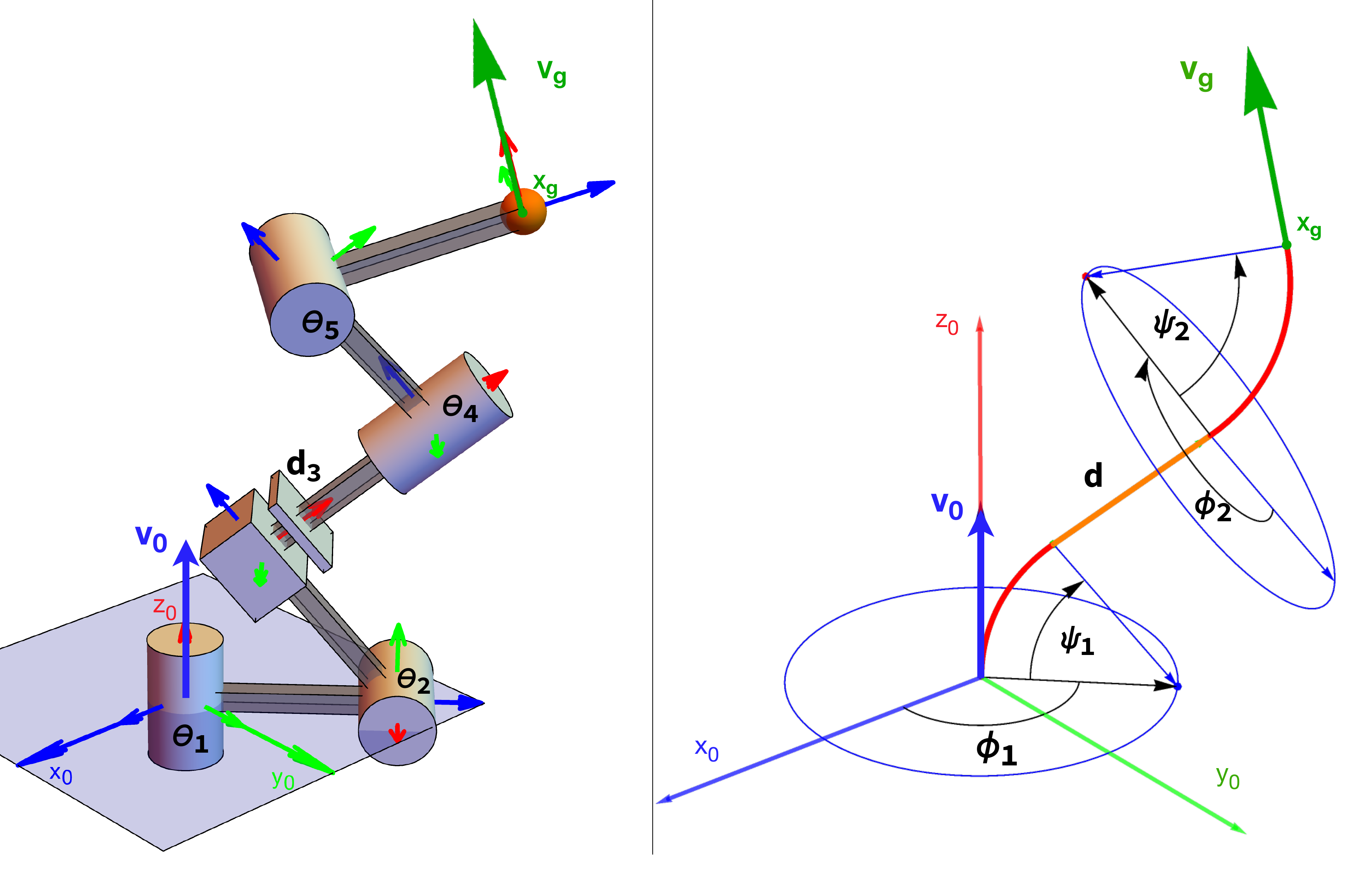}
\caption{\ Figure showing our RRPRR analogue of the 3D CSC Dubins path. The left shows the RRPRR manipulator that we solve the inverse kinematics for, the right shows the equivalent 3D CSC Dubins path. \label{fig:DHtoDubins}
}
\end{figure}

Balkom, Furtuna, and Wang point out a relationship between kinematic descriptions of mobile robots driven by constant controls and serial manipulators in~\cite{balkcom2018DubinsCar}; they conclude this work by suggesting that there may be an arm-like model for 3D Dubins paths.
We model the 3D CSC Dubins path as an RRPRR manipulator, defining $\theta_i$ as the curved components and the prismatic joint as the $d_3$ straight path component. An illustration of our manipulator Dubins path analogue is shown in Fig.~\ref{fig:DHtoDubins}.



\begin{table}[h!]
\centering
\caption{ The CSC Dubins path as a 5-DOF RRPRR arm. \label{tab:DHparamCSC}
}
\vspace{-0.5em}
\begin{tabular}{ |c|c|c|c|c|c| } 
\hline
\multicolumn{6}{| c |}{Denavit-Hartenberg (DH) Parameters}\\
\hline
Joint & Type & $r$ & $\alpha$ & $d$ & $\theta_i$\\
\hline
1 & revolute  & 1 & $\pi /2$ & 0 & $\theta_1$ \\ 
2 & revolute  & 1 & $\pi /2$ & 0 & $\theta_2$ \\ 
3 & prismatic & 0              & 0 & $d_3$ & 0\\ 
4 & revolute  & 1 & $\pi /2$ & 0 & $\theta_4$ \\ 
5 & revolute  & 1 & $\pi /2$ & 0 & $\theta_5$ \\
\hline
\end{tabular}
\end{table}

Table~\ref{tab:DHparamCSC} lists the Denavit-Hartenberg (DH) parameters corresponding to our RRPRR manipulator model shown in Figure~\ref{fig:DHtoDubins}.
DH parameters specify the transformation between consecutive frames as a composition of a rotation about the $z$-axis and a rotation about the rotated $x$-axis~\cite{lynch2017modern}.
 The conversion from DH parameters to Dubins parameters is
\begin{align}
\begin{bmatrix}
\phi_1 \\
\psi_1 \\ 
d \\ 
\phi_2 \\
\psi_2
\end{bmatrix}
=
\begin{bmatrix}
\theta_1 \\
\pi - \theta_2 \\ 
d_3 \\ 
\theta_4 - \pi \\
\pi - \theta_5
\end{bmatrix}.
\end{align}
The relationship between frames $i$ and $i-1$ is
\begin{align}
A_{i} &= 
\begin{bmatrix}
c_{\theta_i} & -s_{\theta_i} c_{\alpha_i} & s_{\theta_i} s_{\alpha_i} & r_i c_{\theta_i}\\
s_{\theta_i} & c_{\theta_i} c_{\alpha_i}  & -c_{\theta_i} s_{\alpha_i} & r_i S_{\theta_i}\\
 0  & s_{\alpha_i} & c_{\alpha_i}& d_i\\ \notag
 0  & 0  & 0 & 1
\end{bmatrix} \\
\end{align}
and the transformation matrix relating the final and first frame is 
\begin{align}
A_1 A_2 A_3 A_4 A_5 &= A_{\textit{hand}}
\label{eq:TransformationMatrix} \, ,
\end{align}
where 
\begin{align}
A_{\textit{hand}} &= 
\begin{bmatrix}
l_x & m_x & v_{g_x} & x_{g_x}\\
l_y & m_y & v_{g_y} & x_{g_y}\\
l_z & m_z & v_{g_z}& x_{g_z}\\
 0 & 0  & 0 & 1 
\end{bmatrix} \, .
\label{eq:Ahand}
\end{align}
The $l$ and $m$ terms are ignored because $\textbf{x}_g$ and $\textbf{v}_g$ only specify the final $z$-axis and position. See Section \ref{sec:SolutionInitialSetup} for details.

\section{solution}
To solve the inverse kinematics problem for our manipulator, we follow closely  the inverse kinematics solution of a 6R manipulator developed by Raghavan and Roth in~\cite{Raghavan1993InverseKinematics}.
The solution involves forming and solving a system of multivariate polynomials in a way that is similar to Gauss elimination but for nonlinear equations -- techniques deeply rooted in resultant-based elimination methods pioneered by Newton and Euler, and later popularized mostly by Sylvester.
The goal is to eliminate all but one variable resulting in a polynomial whose roots can be  substituted back into the system to solve for the remaining variables.
We  use the same notation as~\cite{Raghavan1993InverseKinematics}; any deviation is intended for clarity.
\subsection{Initial setup}\label{sec:SolutionInitialSetup}

We substitute our DH parameters from Table~\ref{tab:DHparamCSC} and begin our solution by first moving the terms in eq.~\eqref{eq:TransformationMatrix} so that
\begin{align}
A_3 A_4 A_5 = A_{2}^{-1} A_1^{-1} A_{\textit{hand}} \, .   
\label{eq:MatrixEquation}
\end{align}

The choice of terms on the left and right of eq.~\eqref{eq:MatrixEquation} is discussed in Section~\ref{subsec:LeftRight}.
From here we work with the six equations from columns three and four of the resulting expansion and refer to them as $\tilde{\textbf{I}}$  and $\tilde{\textbf{p}}$: 

\begin{subequations}
\label{eq:ItildeAndPTilde}
\begin{align}
\tilde{\textbf{I}}=\!\begin{bmatrix}\!
    c_4 s_5 \\
    c_5 \\
    s_4 s_5 \end{bmatrix}
    &\!=\!
\begin{bmatrix}\!
    v_{g_x} c_1 c_2 + v_{g_y} s_1 c_2 + v_{g_z} s_2 \\
    v_{g_z} c_2 - v_{g_x} c_1 s_2 - v_{g_y} s_1 s_2 \\
    -v_{g_y} c_1 + v_{g_x} s_1\!
\end{bmatrix} \label{eq:Itilde} \\
\tilde{\textbf{p}}\!=\!\begin{bmatrix}\!
c_4(1 + c_5) +1\\
-d_3 - s_5\\
s_4(1 + c_5) \!\end{bmatrix}
&\!=\!
\begin{bmatrix}\!
-c_2 ( x_{g_x} c_1 + x_{g_y} s_1 -1) + x_{g_z} s_2\\
s_2(1 - x_{g_x} c_1 - x_{g_y} s_1) + x_{g_z} c_2\\
x_{g_x} s_1 - x_{g_y} c_1\!
\end{bmatrix} \label{eq:pTilde} 
\end{align}
\end{subequations}

We form eight new equations from $\tilde{\textbf{I}}$ and $\tilde{\textbf{p}}$: two equations from $\tilde{\textbf{p}} \cdot \tilde{\textbf{p}} $ and $\tilde{\textbf{p}} \cdot \tilde{\textbf{I}} $; three equations from $\tilde{\textbf{p}} \times \tilde{\textbf{I}} $; and three more equations from $(\tilde{\textbf{p}} \cdot \tilde{\textbf{p}}) \tilde{\textbf{I}} - (2 \tilde{\textbf{p}} \cdot \tilde{\textbf{I}}) \tilde{\textbf{p}}$. 
For reference, see the section, ``The Ideal of $\tilde{\textbf{p}}$ and $\tilde{\textbf{I}}$" in~\cite{Raghavan1993InverseKinematics} or~\cite{IdealTheoryTextBook} on elimination theory in general and proofs regarding the independence of generated ideals. 

These 14 equations are treated as linear combinations of $s_4$ and $c_4$, which ~\cite{Raghavan1993InverseKinematics} calls the ``suppressed" variables. 
Our choice in having the $s_4$ and $c_4$ terms as our suppressed variables and the tradeoffs between choosing these and not the other terms as our suppressed variables is discussed in Section~\ref{subsec:SuppressionChoice}.
The 14 equations are written as
\begin{equation}
\textbf{P}(s_4,c_4)
\begin{bmatrix}
d^2_3 s_5 \\
d^2_3 c_5 \\
d^2_3 \\
d_3 s_5 \\
d_3 c_5 \\
d_3 \\
s_5 \\
c_5 \\
1
\end{bmatrix}
= 
\textbf{Q}(v_{g_i},x_{g_i})
\begin{bmatrix}
s_1 s_2 \\
s_1 c_2 \\ 
c_1 s_2\\ 
c_1 c_2 \\ 
s_1\\ 
c_1 \\ 
s_2\\ 
c_2
\end{bmatrix}
\label{eq:PMat=QMat}\,.
\end{equation}

\subsection[Eliminating theta 1 and theta 2]{Eliminating $\theta_1$ and $\theta_2$}
\label{sec:elim1And2}
Eight out of 14 equations in \eqref{eq:PMat=QMat} are used to solve for the eight terms on the right of \eqref{eq:PMat=QMat} in terms of the eight terms on the left. 
These solutions are substituted into the remaining six equations resulting in a system 

\begin{align}
\label{eq:sigma6}
\boldsymbol{\Sigma} (s_4,c_4,v_{g_i},x_{g_i})
\begin{bmatrix}
d^2_3 s_5 \\
d^2_3 c_5 \\
d^2_3 \\
d_3 s_5 \\
d_3 c_5 \\
d_3 \\
s_5 \\
c_5 \\
1
\end{bmatrix}
= 0_{6\times 1} \, ,   
\end{align}
where $\boldsymbol{\Sigma} (s_4,c_4,v_{g_i},x_{g_i})$ is a $6\times 9$ matrix.

\subsection[Eliminating d3 and theta 5]{Eliminating $d_3$ and $\theta_5$}
\label{sec:elimD3AndTheta5}
Using the tangent half-angle identity, the two terms, $s_4$ and $c_4$ in $\boldsymbol{\Sigma}(s_4,c_4,v_{g_i},x_{g_i})$ can be treated as one variable, $x_4$ with the trigonometric substitution 
\begin{equation}
    s_4 \leftarrow \frac{2 x_4}{1+x_4^2} \quad \text{and} \quad c_4 \leftarrow \frac{1- x_4^2}{1+x_4^2} \nonumber \hspace{2pt} , 
\end{equation}
so that $\boldsymbol{\Sigma}(s_4,c_4,v_{g_i},x_{g_i})$ becomes $\boldsymbol{\Sigma}(x_4, v_{g_i}, x_{g_i})$. 
This $x_4$ can be mapped to a unique $\theta_4$ value.
After multiplying out the denominators, we then multiply each equation in $\boldsymbol{\Sigma}(x_4, v_{g_i}, x_{g_i})$ by $d_3$ to create 6 additional equations (see~\cite{whittaker_1921} and~\cite{EliminationMethods1995} on \emph{resultants} and \emph{dialytic elimination}). 
Although this creates three additional terms ($d^3_3 s_5, d^3_3 c_5, d^3_3$), we arrive at a $12\times12$ system  
\begin{equation}
\begin{bmatrix}
\boldsymbol{\Sigma}(x_4, v_{g_i}, x_{g_i}) & 0_{6\times3 }\\
0_{6\times3 } & \boldsymbol{\Sigma}(x_4, v_{g_i}, x_{g_i}) 
\end{bmatrix}
\begin{bmatrix}
d^3_3 s_5 \\
d^3_3 c_5 \\
d^3_3 \\
d^2_3 s_5 \\
d^2_3 c_5 \\
d^2_3 \\
d_3 s_5 \\
d_3 c_5 \\
d_3 \\
s_5 \\
c_5 \\
1
\end{bmatrix}
= 0_{12\times 1}  \, . 
\label{eq:sigma12}
\end{equation}

To make~\eqref{eq:sigma12} have a non-zero solution, set the determinant 
\begin{equation}
\label{eq:det(diagSig)=0}
\mdet{ \boldsymbol{\Sigma}(x_4, v_{g_i}, x_{g_i}) & 0_{6\times3 }\\
0_{6\times3 } & \boldsymbol{\Sigma}(x_4, v_{g_i}, x_{g_i}) } = 0 \, ,
\end{equation}
and solve~\eqref{eq:det(diagSig)=0} for a single polynomial in terms of $x_4$~\cite{whittaker_1921}. After factoring and removing roots that are always imaginary, we are left with a 12\textsuperscript{th} order polynomial. \cite{Raghavan1993InverseKinematics} calls this the \emph{characteristic polynomial}, because it characterises the maximum number of possible solutions.
Finally we substitute the ${v_{gi}}$ and ${x_{gi}}$ values and solve for the roots. 
\subsection[Solving for theta 4, d3, theta 5 theta 1 and theta 2]{Solving for $\theta_4$, $d_3$, $\theta_5$, $\theta_1$, and $\theta_2$}

For each real root $x_4$, $\theta_4$ is calculated as $\theta_4 = 2 \arctan(x_4)$ and then we proceed with the following two steps to get up to 12 solution sets. 
We discard solution sets with values of $d_3$ that are negative, because these do not apply to the 3D CSC Dubins path.
\subsubsection[d3 and theta 5]{solving for $d_3$ and $\theta_5$}
~\label{subsec:d3AndTheta5}
Substituting the solution for $x_4$ and $v_{g_i}$ and $x_{g_i}$ into 
eq.~\eqref{eq:sigma12} and solving the system yields the solution for $d_3$, and the solutions for the terms $s_5$ and $c_5$. 
Then $\theta_5 = \arctantwo(c_5, s_5)$.

\subsubsection[solving for theta1 and theta2]{solving for $\theta_1$ and $\theta_2$}
~\label{subsec:Theta1AndTheta2}
$v_{g_i}$, $x_{g_i}$ and the solutions for $d_3$, $\theta_4$, and  $\theta_5$ are substituted into the system, eq.~\eqref{eq:PMat=QMat}. Next, eight of the 14 equations are used to solve for the terms in the left column of eq.~\eqref{eq:PMat=QMat}. 
Then $\theta_1 = \arctantwo(c_1,s_1)$ and $\theta_2 = \arctantwo(c_2,s_2)$.

We note here that in Sections \ref{sec:elim1And2} and \ref{sec:elimD3AndTheta5}, when these systems are solved, the multivariate terms, $ d_3^2 s_5 $, $ d_3^2 c_5 $, ... $s_1 s_2$, $s_1 c_2$, ... are treated as single-term variables; this is standard in the treatment of multivariate polynomials using resultant-based elimination methods. 
However, here in step two, the single-term variables for the $d_3$-terms and the $\theta_5$-terms are converted back to multivariate terms when substituting in the solutions for $d_3$, $\theta_4$, and $\theta_5$, while the four multivariate $\theta_1$-and-$\theta_2$-terms are left as single-term variables. 


\section{special cases}
\begin{center}
\textcolor{myDarkGrey}{\emph{``... but as always with geometric algorithms, \\ special cases arise that can be handled separately."}}
\end{center}
\begin{flushright}
\textcolor{myDarkGrey}{\scriptsize{\emph{-- David Eberly \cite{Eberly}}}}
\end{flushright}

There are configurations of $\textbf{x}_g$ and $\textbf{v}_g$ that complicate the system \eqref{eq:sigma12}.
As~\cite{EliminationMethods1995} points out, methods that extract resultants from determinants fail for singular matrices. 
For configurations that make the coefficient matrix in eq.~\eqref{eq:det(diagSig)=0} singular, we cannot generate a characteristic polynomial because the determinant is always zero. 
Fortunately, a few algebraic and geometric observations allow us to find solutions for these configurations.

\subsection{Configurations with one solution}
If $\textbf{v}_g$, $\textbf{x}_g-\textbf{x}_0$, and $\textbf{v}_0$ are parallel, then the goal configuration is reached with a straight line path.

\subsection{Configurations with infinite solutions}
\label{subsec:InfiniteSolutions}
\begin{figure}[tb]
\centering
\includegraphics[width=1\columnwidth]
{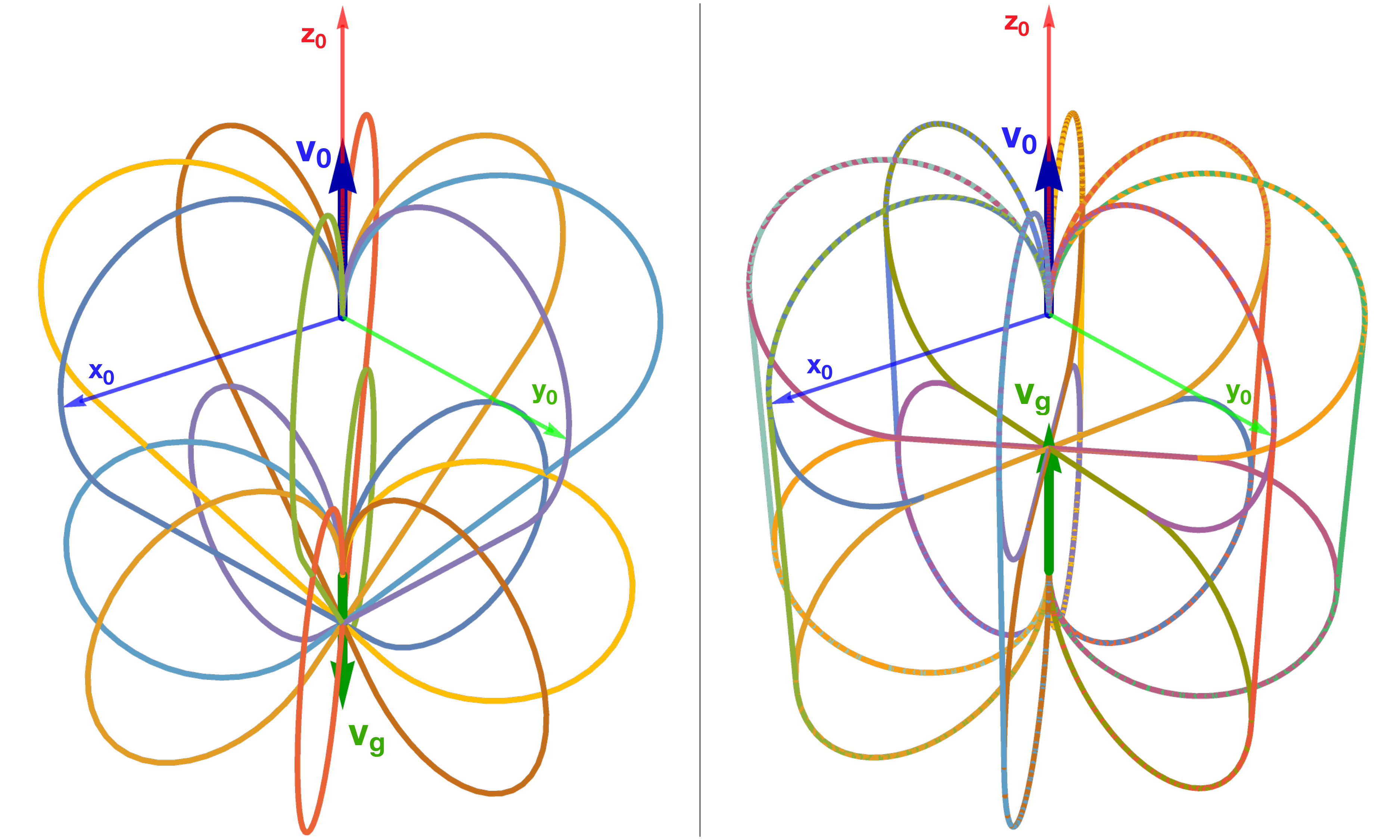}
\caption{An example of configurations with infinite valid CSC Dubins paths shown with representative paths. The goal orientation $\textbf{v}_g$  and the goal position $\textbf{x}_g$ are colinear with $\textbf{v}_0$ and $\textbf{x}_0$. On the left, $\textbf{v}_g = [0,0,-1]$ and on the right, $\textbf{v}_g = [0,0,1]$. 
\label{fig:InfiniteSolutionExample}
}
\end{figure}

If both $\textbf{v}_g$ and $\textbf{x}_g$ are aligned with $\textbf{v}_0$, then there will be infinite valid paths. 
The first arc begins in any direction, $\theta_1$, the length of the straight segment will depend on the distance $\textbf{x}_g - \textbf{x}_0$, and the second arc will be on the same plane as the first so that the final orientation is colinear with the first. 
Since we know where the second arc will be, we can supply $\theta_1$ and $\theta_4$ pairs to equations that have not vanished in system~\eqref{eq:PMat=QMat} and solve for $\theta_2$, $d_3$, and $\theta_5$.
If there is sufficient distance, $\textbf{x}_g - \textbf{x}_0 > 2r$, and $\textbf{v}_g$ is in the same direction as $\textbf{v}_0$, then there will be be two $\theta_4$-solutions for any $\theta_1$ direction ($\theta_4 = \left\{{0,\pi}\right\}$), but when $\textbf{v}_g$ and $\textbf{v}_0$ are in the opposite directions, and $\textbf{x}_g - \textbf{x}_0 \le 2r$, then there will only be one $\theta_4$-solution for any $\theta_1$ direction ($\theta_4 = 0$). 
This is illustrated in Figure~\ref{fig:InfiniteSolutionExample} for both directions of $\textbf{v}_g$ at the distance threshold, $\textbf{x}_g - \textbf{x}_0 = 2r$. Both images show sample paths with eight supplied $\theta_1$ values. The image on the right has 16 corresponding paths when $\textbf{v}_g$ and $\textbf{v}_0$ are in the same direction, while the image on the left has eight paths because $\textbf{v}_g$ and $\textbf{v}_0$ are in opposite directions.

\subsection{Planar paths in 3D}
\label{subsec:PlanarPaths}

If $\textbf{v}_g$ is in the plane that contains the initial orientation $[0,0,1]^\top$ and $\textbf{x}_g$, then the problem is 2D, and can be solved using the methods of \cite{dubins1957curves}.
In these cases, we can show that our model suggests planar paths even in 3D space.

In Section \ref{sec:elim1And2}, when forming $\boldsymbol{\Sigma}(s_4,c_4,v_{g_i},x_{g_i})$, many of the terms contain a constant denominator, $x_{g_x} v_{g_y} - x_{g_y} v_{g_x}$. 
This quantity is the $z$-component of $\textbf{x}_g \times \textbf{v}_g$. 
Naturally we multiply this quantity out of the denominators when forming the systems \eqref{eq:sigma6}, \eqref{eq:sigma12}, and eq.~\eqref{eq:det(diagSig)=0} for the general solution, but the multiples are still present in the system. 
When $\textbf{x}_g \times \textbf{v}_g$ is on the $x_0y_0$-plane, $x_{g_x} v_{g_y} - x_{g_y} v_{g_x} = 0$, and system~\eqref{eq:sigma12} loses many equations, so we cannot use it to generate a characteristic polynomial or solve for $d_3$ and $\theta_5$.

In Section~\ref{sec:elim1And2}, system~\eqref{eq:sigma6} is in the form 
$\textbf{A\,b} = 0$ where $\textbf{A}$ is a $6 \times 8$ matrix and $\textbf{b}$ is an $8 \times 1$ column vector. 
After applying the condition that $x_{g_x} v_{g_y} = x_{g_y} v_{g_x}$ to  system~\eqref{eq:sigma6}, $\textbf{A}$ is a $n \times m$ matrix and $\textbf{b}$ is an $m \times 1$ column vector where $n<6$ and $m<8$. 
This system can be expanded dialytically to generate the singular  $m \times n$ matrix $\textbf{A}$, where $m=n<12$. 
The  resultant characteristic polynomial's factors reveal that the roots are all either imaginary or zero.

\begin{figure}[tb]
\centering
\includegraphics[width=1\columnwidth]
{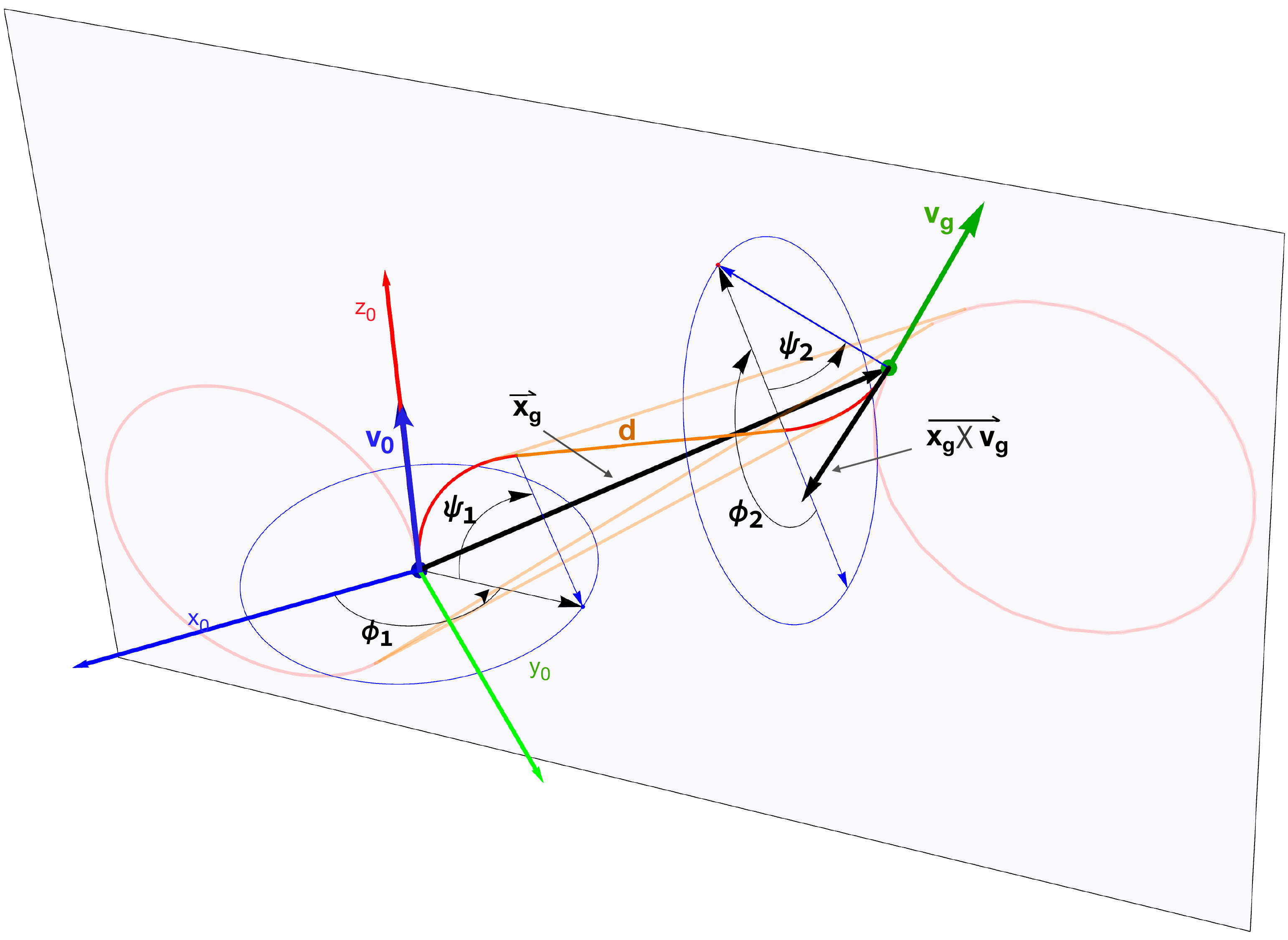}
\caption{Planar paths in 3D occur when $\textbf{x}_g \times \textbf{v}_g$ is only in the $x_0y_0$ plane.  
The four valid CSC paths in this figure are all on the light blue plane defined by $\textbf{v}_0$, $\textbf{v}_g$, and $\textbf{x}_g$.
The blue circles are orthogonal to this plane and show the reachable orientations of the $\psi_1$ and $\psi_2$ arcs. 
\label{fig:PlanarCase}}
\end{figure}

Now that we know $x_4$ can only equal zero for these configurations, we can use its geometric implications to find $\theta_1$, and our original system to solve for $\theta_2$, $d_3$, and $\theta_5$. 
The remainder of this subsection refers to Fig.~\ref{fig:PlanarCase}, which shows an example of a CSC configuration with planar-path solutions, and we use Dubins-path notation mapped from DH parameters through eq.~\eqref{fig:DHtoDubins}.

To start, we know that $x_4 = 0$ when $\theta_4$ (or $\phi_2$) is either $0$ or multiples of $\pi$, defining one possible plane for the 
$r \psi_2$ arc. 
If we know that the 
$r \psi_2$ arc
is on the plane,
$\phi_2 = \left\{{0,\pi}\right\}$,
then we can also conclude that arc
$r \psi_1$
must be on the same plane, because that is the only way that it can produce a trajectory that will arrive on this 
\mbox{$\phi_2$-plane}.
Although a single 
$\phi_1$-plane
can define infinite 
\mbox{$\phi_2$-planes}
in the forward kinematics problem, a fixed 
$\phi_2$-plane
governs only one possible 
$\phi_1$-plane
in the inverse kinematics problem.
Using this logic, we can compute 
$\phi_1$
from the projection of the position vector  $\textbf{x}_g$ onto the $xy$ plane,
$\phi_1 = \arctantwo(x_{g_x},x_{g_y})$. 
But this is also the same plane as 
$\phi_1 = \arctantwo(x_{g_x},x_{g_y}) + \pi.$  
So for both 
$\phi_2 = \left\{{0,\pi}\right\}$, and for both $\phi_1 =\left\{{\arctantwo(x_{g_x},x_{g_y}), \arctantwo(x_{g_x},x_{g_y}) + \pi}\right\}$ (or $\phi_1 =\left\{{\arctantwo(v_{g_x},v_{g_y}), \arctantwo(v_{g_x},v_{g_y}) + \pi}\right\}$ if $ x_{g_y}=x_{g_x}=0$) mapped to $\theta_1$ and $\theta_4$, we can use parts of the original system~\eqref{eq:PMat=QMat} to solve for $\theta_2$, $d_3$, and $\theta_5$ for a total of four solution sets; this is the order of the repeating zero-roots of the characteristic polynomial. 
Two equations from  $\tilde{\textbf{p}}$, and two equations from $\tilde{\textbf{I}}$ are  used to solve for a unique $\theta_2$ and $\theta_5$ in terms of $d_3$, and then substituted into $\tilde{\textbf{p}} \cdot \tilde{\textbf{p}}$ to solve for $d_3$.
Substituting $\theta_2$ and $\theta_5$ as functions of $d_3$ into any of the remaining equations from system~\eqref{eq:PMat=QMat} that have not vanished from the condition $x_{g_x} v_{g_y} = x_{g_y} v_{g_x}$ result in a $d_3$-polynomial of order $n \in \{2,3,4\} $, and the real roots of these $d_3$-polynomials are all  equal.

There are additional singular cases that need to be handled separately, but due to page limitations we will document them in future work.


\section{results}

\begin{figure}[tb]
\centering
\includegraphics[width=1\columnwidth]{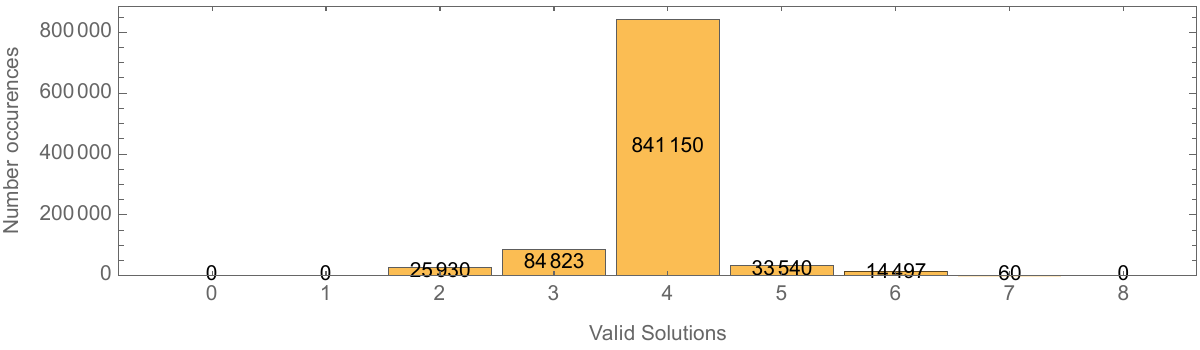}
\caption{Number of valid CSC paths for one million trials, with $\textbf{x}_g$ uniformly sampled in $[-4,4]^3$ and $\textbf{v}_g$ uniformly sampled on the unit sphere.} 
\label{fig:FreqSolutionsDubins3D}
\end{figure}
\begin{table}[]
\begin{tabular}{|r rrrr rrrr r|}
\hline
\!\!\!solutions:\! \!& \!\!0 & \!1 \! & \!2     & \!3    & \!4     & \!5    & \!6    &\!7 \!     &\!\! 8 \!\!\!\\
\!\!\% found:  \!\!& \!\!0 & \!0 \! & \!2.59  & \!8.48 & \!84.1 & \!3.35 & \!1.44 & \!0.006\! & \!\!0 \!\!\!\\
\hline
\end{tabular}
\caption{\label{tab:numSolutions}  CSC solution distribution for one million trials }
\end{table}

\begin{figure}[tb]
\centering
\includegraphics[width=1\columnwidth]
{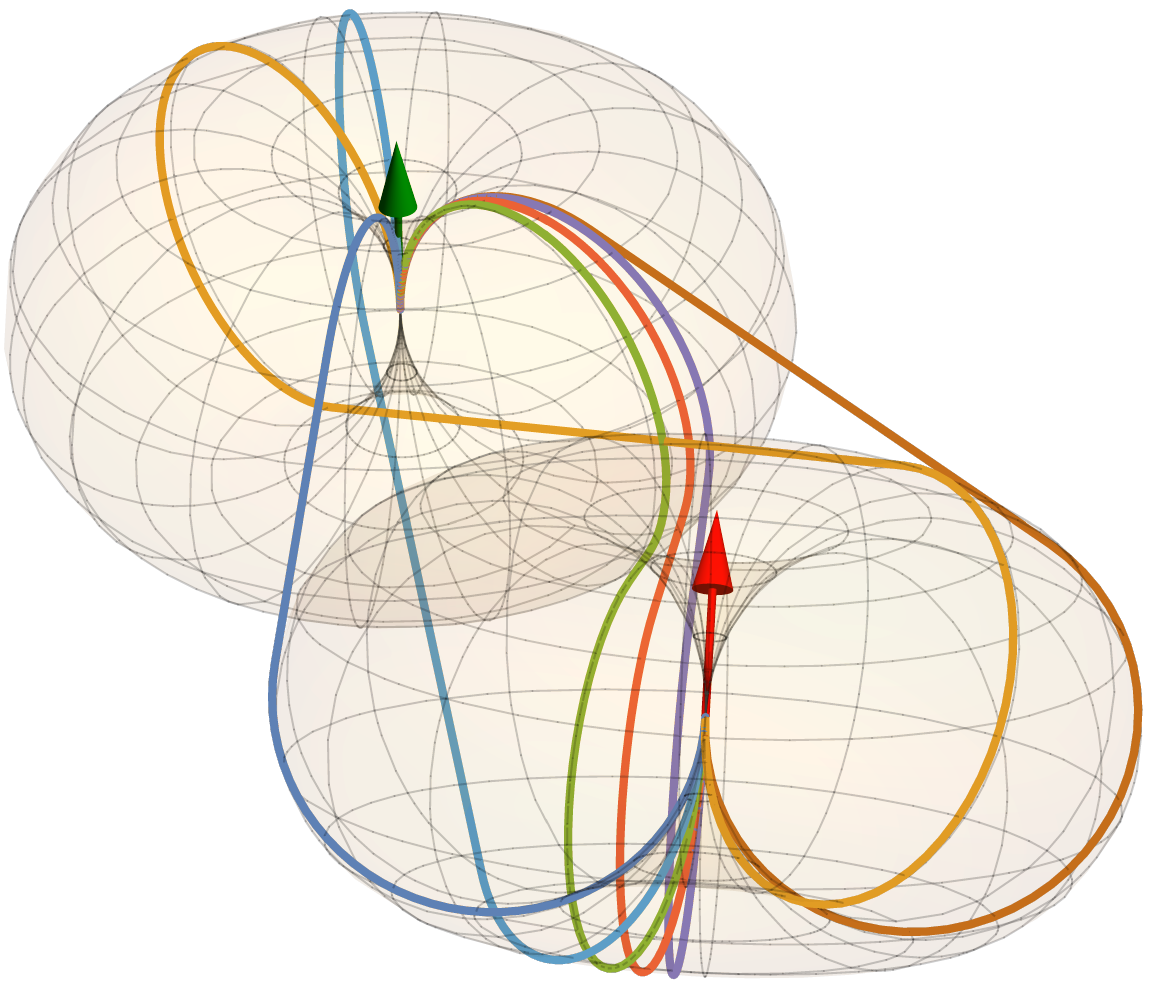}
\caption{A configuration with seven valid CSC Dubins paths in 3D. The initial position and orientation at $\textbf{x}_0 = [0,0,0]^\top$, $\textbf{v}_0 = [0,0,1]^\top$ is shown with a green arrow.
The goal position and goal orientation $\textbf{x}_g = [2.64101, -1.78042, -0.371051]^\top$, $\textbf{v}_g = [-0.323321, 0.729589, 0.602631]^\top$ are shown with a red arrow.
The tori represent the minimum turning radius $r=1$.} 
\label{fig:7SolutionExample1}
\end{figure}

To examine the variation in the number of CSC solutions over the configuration space, we generated one million sample points uniformly in the $[-4,4]$ cube and $\textbf{v}_g$  uniformly sampled on the the unit sphere~\cite{marsaglia1972choosing}. We then solved the inverse kinematics. A histogram of the number of solutions found are shown in Fig.~\ref{fig:FreqSolutionsDubins3D}, and in Table \ref{tab:numSolutions}.
Each inverse kinematics query required on average 0.025 seconds. 
For 84\% of configurations four solutions were found.
The inverse kinematics always found at least two solutions. 
In 88.9\% of the configurations at least four solutions are found, and seven valid solutions were found 60 times. 
One example of a configuration with seven valid CSC paths is shown in Fig.~\ref{fig:7SolutionExample1}.

\begin{figure}[t]
\centering
\includegraphics[height=0.53\columnwidth,trim={.05cm 0 0.05cm 0},clip]
{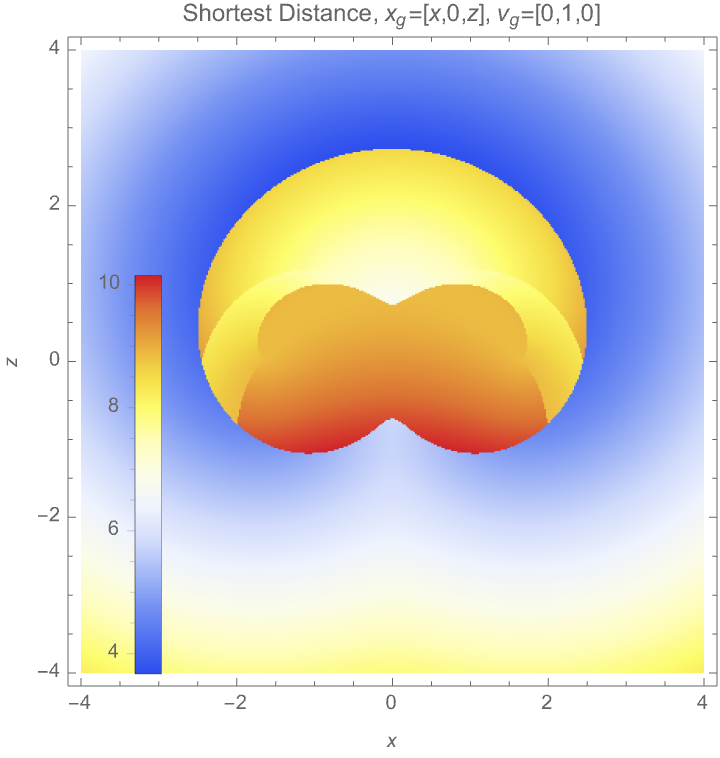}
\includegraphics[height=0.53\columnwidth,trim={.5cm 0 0.04cm 0},clip]
{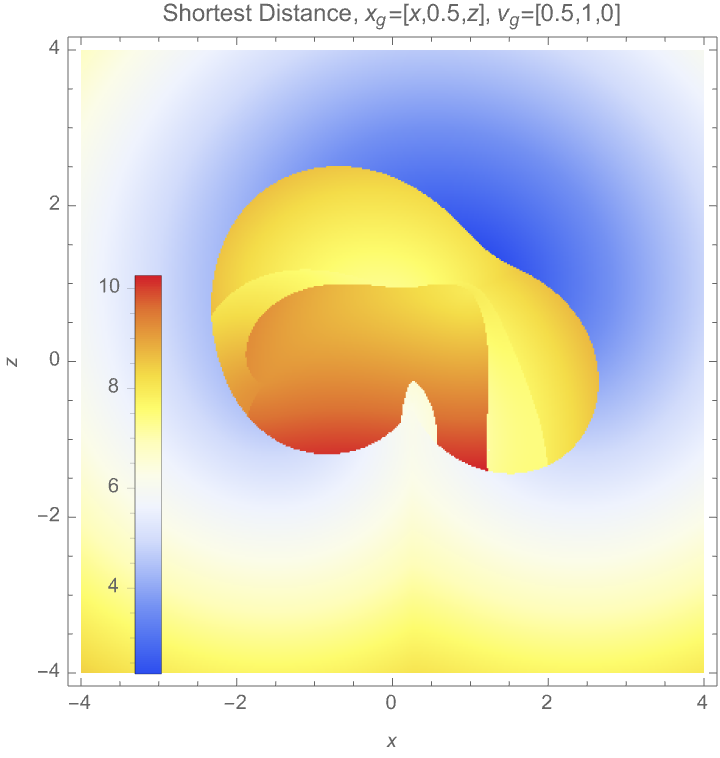}
\caption{
2D slices of the configuration spaces in Fig.~\ref{fig:2DsliceNumberSols}, showing the length of the shortest solution to the CSC Dubins path for each configuration. }
\label{fig:2DsliceDist}
\end{figure}

Just as the Dubins 2D solution is not a smooth distance function and is not a metric, the distance returned by the shortest 3D CSC paths is also not smooth.
  The shortest CSC distance plotted for two slices of the configuration space is shown in Fig.~\ref{fig:2DsliceDist}. 
  The plot for  $\textbf{x}_g = [x,0,z]^\top$, $\textbf{v}_g = [0,1,0]^\top$ is relatively simple. Outside the cardioid-shaped region at the center the shortest CSC distance increases with distance from the origin.
  However, CSC paths inside the cardioid-shaped region have large lengths, especially along the bottom edge. 
  The plot for $\textbf{x}_g = [x,0.5,z]^\top$, $\textbf{v}_g = [0.5,1,0]^\top$ is more complicated, with multiple regimes and a long-distance c-shaped region at $[x,z] = [-2,-0.25]$.

\begin{figure}[tb]
\centering
\includegraphics[width=1\columnwidth]
{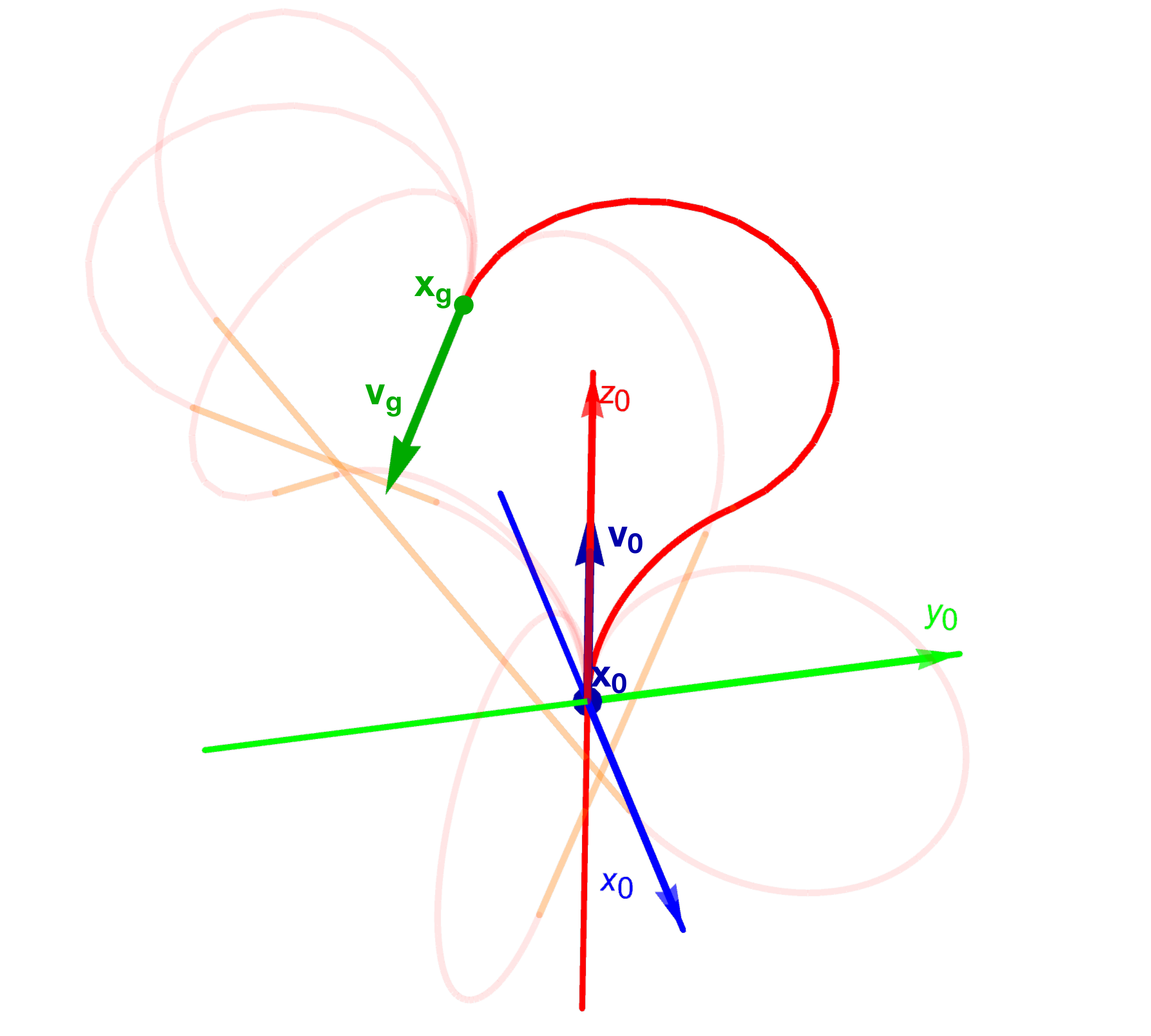}
\caption{A configuration with five solutions, where one of the valid paths (highlighted in red) has a zero-length straight line component. 
\label{fig:d3IsZero}}
\end{figure}

Our solution is also robust to configurations where the straight line component has zero length. An example of a configuration with five solutions where one of them has a component $d=0$ is shown in Fig.~\ref{fig:d3IsZero}.

\section{Discussion}
\label{sec:Discussion}
Critics of dialytic elimination methods warn about the potential for explosive growth in the order of the resultant polynomial for systems where more than one multiple is needed during the expansion process to arrive at a square matrix~\cite{EliminationMethods1995,diankov2010automated}.
For reference, our expansion for the general solution is in Section~\ref{sec:elimD3AndTheta5}.
In this section we discuss how we avoid such explosive growth and other particularities used in our solution. 

\subsection{Choice of left and right terms in the matrix model}
\label{subsec:LeftRight}

The combination of terms on the left and right of \eqref{eq:MatrixEquation} is intentional. 
This combination results in a system of the form:
\begin{equation}\small
\begin{bmatrix}
f(\theta_{4,5}) & f(\theta_4) & f(\theta_{4,5}) &f(\theta_{4,5})\\
f(\theta_5) & 0 & f(\theta_5) & f(d_3,\theta_5)\\
f(\theta_{4,5})& f(\theta_4) & f(\theta_{4,5})& f(\theta_{4,5})\\
 0 & 0  & 0 & 1\\ 
 \end{bmatrix} \nonumber
  =\\
\begin{bmatrix}\small
 f(\theta_{1,2},l) & f(\theta_{1,2},m) & f(\theta_{1,2},v_{g}) &f(\theta_{1,2},x_{g})\\
f(\theta_{1,2},l) & f(\theta_{1,2},m) & f(\theta_{1,2},v_{g}) & f(\theta_{1,2},x_{g})\\
f(\theta_1,l)& f(\theta_1,m) & f(\theta_1,v_{g})& f(\theta_1,x_{g})\\
 0 & 0  & 0 & 1\\ 
\end{bmatrix} 
\label{eq:SystemForm}
\end{equation} 
Different combinations result in different systems, some of which seem more simple because they yield equations with less variables to solve for. 
For example, instead of \eqref{eq:MatrixEquation}, we could formulate the system as 
\begin{align}
A_3 A_4  = A_{2}^{-1} A_1^{-1} A_{\textit{hand}} A_5^{-1}.
\end{align}
which would expand to a system of the form: 
\begin{equation}\small
\label{eq:SystemForm2}
\begin{bmatrix}
f(\theta_4) & 0 & f(\theta_4) &f(\theta_4)\\
0 & -1 & 0 & f(d_3)\\
f(\theta_4)& 0 & f(\theta_4)& f(\theta_4)\\
 0 & 0  & 0 & 1\\ 
 \end{bmatrix} 
  =\\
\begin{bmatrix}\small
 f(\theta_{1,2,5},l,v_{g}) &  f(\theta_{1,2,5},l,v_{g}) & f(\theta_{1,2},m) & f(\theta_{1,2},l,x_{g})\\
 f(\theta_{1,2,5},l,v_{g}) & f(\theta_{1,2,5},l,v_{g}) & f(\theta_{1,2},m) & f(\theta_{1,2},l,x_{g})\\
f(\theta_{1,5},l,v_{g})& f(\theta_{1,5},l,v_{g}) & f(\theta_1,m)& f(\theta_{1,5},l,x_{g})\\
 0 & 0  & 0 & 1\\
\end{bmatrix} \nonumber
\end{equation} 
Comparing the system in~\eqref{eq:SystemForm} to~\eqref{eq:SystemForm2}, the columns 3 and 4 do not contain $\theta_5$ in~\eqref{eq:SystemForm2}, while they do in~\eqref{eq:SystemForm}. The system~\eqref{eq:SystemForm2} would require less variables to eliminate, and then $\theta_5$ could have been solved for by substituting the 1 through 4 solutions into 
\begin{align}
  A_5= A_4^{-1}A_3^{-1}A_{2}^{-1} A_1^{-1} A_{\textit{hand}}.
\label{eq:MatrixEquationSolveA_5}
\end{align}
However, the combination in \eqref{eq:MatrixEquation} accomplishes two things.
1.) For some combinations, when eliminating the first set of variables as in Section~\ref{sec:elim1And2}, we would be left with less independent equations in system~\eqref{eq:sigma6} requiring us to expand more than once to match the number of equations to the number of extra terms introduced to the system. 
2.) This combination leaves our six equations from columns 3 and 4 in terms of only the last two columns of $A_{\textit{hand}}$, $x_{g_i}$ and $v_{g_i}$. 
Other combinations of eq.~\eqref{eq:MatrixEquation} result in a system such as~\eqref{eq:SystemForm2}, where all the the columns that contain the $x_{g_i}$ and $v_{g_i}$ constants also contain the $l_i$ and $m_i$ constants.
Using our form, we are able to keep our solution as a function of only the goal position, {$\textbf{x}_g$} and goal orientation, $\textbf{v}_g$ and we do not have to create compatible $l$ and $m$ columns of $A_{\textit{hand}}$ to solve our system. 

\subsection{Choice of suppressed variable}
\label{subsec:SuppressionChoice}
Our manipulator's DH parameters required the suppression of the $\theta_4$ terms. 
If for example, we suppressed $d_3$ and the system was treated as linear combinations of the $d_3$-terms instead of the $\theta_4$ terms, the resulting $12\times12$ system as in eq.~\eqref{eq:sigma12} would suffer from column dependencies that could be traced back to the coefficients of the $s_4$ and $s_4 c_5$ terms being equal --- that and the fact that at every instance, they came in a pair. This would also require us to expand more than once, leading to a larger system.

\subsection{Careful expansion for planar paths}
\label{subsec:CarefulExpansion}
Section~\ref{subsec:PlanarPaths} mentions  we can arrive at an $n \times m$ matrix $\textbf{A}$ where $n = m \le 12$ to begin the solution for the special case of planar paths in 3D.
After applying the condition for this case, our system~\eqref{eq:PMat=QMat} reduces, but it still behaves in our favor and allows us to avoid explosive growth.
By expanding dialytically on only the equations that remain after eliminating the first set of variables, we avoid introducing extra terms and still arrive at a square matrix without requiring more than one expansion. 
 
\section{Conclusions}

This paper described a method to determine the valid CSC paths for a Dubins car model in 3D.
This model obeyed a minimum turning radius of $r$.  
The solution was generated by describing the trajectory as an RRPRR robot arm.  
Self-collision of the arm on this trajectory was not accounted for, but established methods for robot arms could be applied~\cite{rakita2018relaxedik}.
The work of Chen et al.\ had joint limits for the generated CSC robot arms~\cite{chen2022kinegami}.  Since our method returns all solutions, they can be easily filtered using axis joint limits to return suitable paths. 
These methods may also apply to closed-form inverse kinematic solutions for concentric tube robots~\cite{dupont2009design,mitros2022theoretical}.
An interesting direction for future work would be to expand this method to generate CCC shortest paths, along with Sussman's  helicoidal paths~\cite{Sussmann1995CDC}, to determine the true shortest paths. 
\balance
\bibliography{bib}
\bibliographystyle{IEEEtran}

\end{document}